\documentclass[letterpaper]{article}

\usepackage[preprint]{aaai2027}

\usepackage[para]{footmisc}
\makeatletter

\newcommand{\projectleader}{%
  \footnotemark[2]%
}

\newcommand{\projectleadernote}{%
  \ifx\thanks\relax
  \else
    \stepcounter{footnote}%
    \protected@xdef\@thanks{%
      \@thanks
      \protect\footnotetext[\the\c@footnote]{Project leader.}%
    }%
  \fi
}

\makeatother

\usepackage[hyphens]{url}
\usepackage{graphicx}
\usepackage{natbib}
\usepackage{caption}
\usepackage{algorithm}
\usepackage{algorithmic}

\usepackage{newfloat}
\usepackage{listings}
\DeclareCaptionStyle{ruled}{
  labelfont=normalfont,
  labelsep=colon,
  strut=off
}
\floatstyle{ruled}
\newfloat{listing}{tb}{lst}{}
\floatname{listing}{Listing}

\usepackage{booktabs}

\usepackage{amsmath}
\usepackage{amssymb}
\usepackage{multirow}

\title{
TennisVAR: A Stroke-Evidence-Grounded Multimodal Large Language Model\\
for Tactical Reasoning in Tennis Videos
}

\author{
Yifan Mei\textsuperscript{\rm 1},
Qinglin Shi\textsuperscript{\rm 1},
Changli Wu\textsuperscript{\rm 1,\rm 2}\projectleader,
Jiayuan Rao\textsuperscript{\rm 3},
Jiayi Ji\textsuperscript{\rm 1},
Liujuan Cao\textsuperscript{\rm 1}\corresponding
\projectleadernote
}

\affiliations{
\textsuperscript{\rm 1}
Key Laboratory of Multimedia Trusted Perception and Efficient Computing,\\
Ministry of Education of China,
Xiamen University\\
\textsuperscript{\rm 2}
Shanghai Innovation Institute\\
\textsuperscript{\rm 3}
School of Artificial Intelligence,
Shanghai Jiao Tong University
}

\begin{document}

\maketitle

\begin{abstract}
Sports-video understanding is moving beyond event recognition toward explaining how actions collectively shape match progression. However, existing tennis-video methods either perceive individual strokes without modeling their tactical dependencies or generate high-level analyses without grounding them in the underlying events. To bridge this perception-to-understanding gap, we formulate \textbf{stroke-evidence-grounded tactical reasoning}, a new rally-level task that requires models to jointly predict an open-ended answer, a hierarchical tactic label, an ordered sequence of supporting strokes, and decisive key actions, with each evidence stroke anchored to its racket--ball contact frame. 
We further introduce \textbf{TRACE} (\textbf{T}actical \textbf{R}easoning with \textbf{A}ction-\textbf{C}hain \textbf{E}vidence in Tennis), a large-scale expert-annotated benchmark containing 11,189 rally videos, 41,485 stroke events, 25,429 tactical units, and 11,189 question--answer pairs. TRACE unifies fine-grained stroke attributes, cross-stroke tactical relations, hierarchical tactic annotations, and evidence-grounded questions across factual perception, tactical understanding, and decision reasoning. 
Building on TRACE, we propose \textbf{TennisVAR} (Tennis \textbf{V}ideo \textbf{A}ction-chain \textbf{R}easoner), an evidence-grounded multimodal large language model that follows an ``event--relation--evidence--tactic'' reasoning paradigm. An Event Parsing Module converts continuous rallies into explicit stroke-event sequences, while a Tactical Graph-Guided Temporal Reasoner jointly models rally progression and same-player decision dependencies to identify question-relevant evidence and decisive actions. 
TennisVAR achieves 73.04 T-F1@8, 56.19 T-IoU@4, and 70.98 hierarchical tactic F1, outperforming the strongest supervised baselines by 19.94, 33.03, and 6.08 points, respectively.
More importantly, it substantially improves the localization and attribution of stroke-level evidence supporting its predictions.

\textbf{Project page:} \url{https://whynotgit2025.github.io/TennisVAR/}.

\end{abstract}


\section{Introduction}
\label{sec:introduction}

Sports video understanding is evolving from action recognition and event localization toward rally-level description, relational reasoning, and tactical analysis~\cite{deliege2021soccernet,shao2020finegym}. Advances in multimodal large language models (MLLMs)~\cite{zhang2023videollama,maaz2024videochatgpt,bai2025qwen3vl} have substantially improved event-level understanding. Yet understanding a sports match requires more than recognizing individual actions; it also requires explaining how those actions interact to shape the progression of play. This is particularly important in tennis, where the natural unit of understanding is an entire rally rather than an isolated stroke.

Existing tennis-video research has progressed along two largely separate directions. Ball-tracking methods and fine-grained event benchmarks can localize racket--ball contacts and recognize attributes such as the hitter, stroke type, direction, and outcome~\cite{huang2019tracknet,liu2025f3set}. They provide precise event-level perception, but largely treat strokes independently and cannot explain their tactical interactions. Recent video-language models instead represent rallies as ordered stroke sequences and generate professional commentary and analysis~\cite{li2026sportsqa,xia2025sportu,rao2025unisoccer,xia2026sportr}, enabling higher-level semantic understanding. However, their predictions may rely on rally outcomes or language priors without explicit grounding in the underlying events. This creates a fundamental \textbf{perception-to-understanding gap}: existing methods neither reconstruct how earlier strokes shape subsequent decisions nor identify the specific strokes supporting a tactical conclusion, making it difficult to verify whether an analysis truly reflects the rally process.

To bridge this gap, we formulate \textbf{stroke-evidence-grounded tactical reasoning}, which requires models to derive tactical conclusions from the specific stroke events that support them. Given a rally video and a natural-language question, a model must jointly predict an open-ended answer, a hierarchical tactical label, an ordered sequence of supporting strokes, and a subset of decisive key actions, with each evidence stroke anchored to its corresponding racket--ball contact frame. The task therefore evaluates not only whether a tactical answer is correct, but also whether the model can reconstruct and ground the reasoning process behind it. 

To support this task, we construct \textbf{TRACE} (\textbf{T}actical \textbf{R}easoning with \textbf{A}ction-\textbf{C}hain \textbf{E}vidence in Tennis), a large-scale, expert-annotated benchmark for rally-level tactical reasoning. TRACE contains 11,189 rally videos, 41,485 stroke events, 25,429 tactical units, and 11,189 question--answer pairs. Each rally is annotated with fine-grained stroke attributes, cross-stroke tactical relations, and explicit links between tactical answers and their supporting evidence. TRACE further introduces a hierarchical tactic ontology with 6/17/25 classes and organizes its questions and evidence chains into three progressive reasoning levels: \emph{factual perception}, \emph{tactical understanding}, and \emph{decision reasoning}. By unifying stroke perception, relational reasoning, tactical prediction, and evidence attribution, TRACE provides a systematic test of whether a model can move from recognizing individual events to understanding the tactical progression of an entire rally.

General-purpose MLLMs remain limited in rally-level tactical reasoning. While they can recognize individual strokes and generate fluent descriptions, they often fail to organize temporally distributed events into a coherent tactical chain, distinguish setup strokes from decisive actions and outcomes, or identify the evidence that supports a tactical conclusion~\cite{fu2025videomme,wu2024longvideobench,xiao2024nextgqa}. The central challenge is therefore not merely recognizing \emph{what happened}, but modeling how strokes functionally depend on one another and jointly shape the rally. Without explicit relational structures and supervised evidence selection, MLLMs may generate plausible analyses that are weakly grounded in the actual match process.

To address this challenge, we propose \textbf{TennisVAR} (Tennis \textbf{V}ideo \textbf{A}ction-chain \textbf{R}easoner), an evidence-grounded MLLM that performs structured reasoning from stroke events to tactical conclusions. TennisVAR first introduces an \emph{Event Parsing Module (EPM)} that converts a continuous rally video into an ordered sequence of semantically explicit stroke events, providing discrete and interpretable primitives for reasoning. It then employs a \emph{Tactical Graph-Guided Temporal Reasoner (TGTR)}, which constructs a typed graph over these events and jointly models two complementary dependencies: the temporal progression between consecutive strokes and the same player's decision transitions across intervening opponent returns. Conditioned on the question, TGTR identifies supporting evidence and decisive actions and integrates them for hierarchical tactical prediction and answer generation. This structured ``event--relation--evidence--tactic'' paradigm explicitly reconstructs how a tactic unfolds across strokes, making each tactical conclusion traceable to the rally events that support it.

Our main contributions are threefold:
\begin{itemize}
\item We formulate \textbf{stroke-evidence-grounded tactical reasoning}, a new task that jointly predicts tactical answers, hierarchical labels, supporting strokes, and decisive actions, extending sports-video understanding from event recognition to evidence-grounded tactical reasoning.

\item We introduce \textbf{TRACE}, the first large-scale expert-annotated benchmark that unifies fine-grained stroke events, cross-stroke tactical relations, hierarchical tactics, and evidence-grounded question answering.

\item We propose \textbf{TennisVAR}, an evidence-grounded MLLM with an ``event--relation--evidence--tactic'' paradigm. It explicitly parses stroke events and models rally progression and same-player decision dependencies for traceable tactical prediction.
\end{itemize}

\section{Related Work}
\label{sec:related_work}

\subsection{Fine-Grained Sports Video Understanding}

Sports benchmarks increasingly expose fine-grained temporal and semantic structure. SoccerNet-v2 supports action spotting and replay grounding in broadcast soccer~\cite{deliege2021soccernet}; FineGym and FineDiving decompose complex routines into structured actions or phases~\cite{shao2020finegym,xu2022finediving}; and F$^3$Set provides dense timestamps for fast, frequent events, including tennis strokes~\cite{liu2025f3set}. Domain-specific vision--language work further addresses soccer understanding and commentary generation~\cite{rao2025unisoccer,rao2024matchtime}.
These studies establish structured events as an important basis for sports understanding, but primarily evaluate event predictions or generated descriptions. Our work instead evaluates whether a high-level tactical judgment is supported by the relevant domain events.

\subsection{Tennis Video Understanding and Tactical Analysis}

Tennis analysis has progressed from ball tracking to structured rally modeling.
TrackNet estimates fast ball trajectories with heatmap representations~\cite{huang2019tracknet}, and F$^3$Set provides precise contact timestamps and compositional stroke labels~\cite{liu2025f3set}.
TennisTV evaluates MLLMs on ordered stroke sequences at stroke and rally levels~\cite{bao2025tennistv}, while TennisExpert combines structured parsing with hierarchical temporal memory for analytical commentary~\cite{liu2026tennisexpert}.
These methods improve perception and domain-specific generation, but do not treat the supporting strokes of each tactical conclusion as an explicit prediction target.
TRACE associates each tactical answer with semantically indexed evidence strokes, their contact frames, and a subset of decisive key actions. Models must therefore predict both the answer and the ordered strokes that justify it, enabling direct evaluation of event--tactic reasoning.

\subsection{Evidence-Grounded Video Reasoning}
General-purpose video-language models enable open-ended interaction with video content~\cite{zhang2023videollama,maaz2024videochatgpt,bai2025qwen3vl,zhou2025glimpse}, while benchmarks such as Video-MME and LongVideoBench evaluate temporal reasoning over extended videos~\cite{fu2025videomme,wu2024longvideobench}. Grounded VideoQA further examines whether answers are supported by relevant evidence: NExT-QA and NExT-GQA study causal and temporal reasoning~\cite{xiao2021nextqa,xiao2024nextgqa}, MMR-V considers multiple temporally distributed evidence segments~\cite{zhu2026mmrv}, and CaST-Bench evaluates multi-evidence causal chains~\cite{zhang2026cast}. These works demonstrate that answer correctness alone is insufficient for reliable video reasoning.
Unlike generic temporal grounding, TRACE represents evidence as chronologically ordered stroke events with explicit tennis semantics. It jointly evaluates hierarchical tactical prediction, evidence-stroke localization, key-action identification, and contact-frame accuracy. To our knowledge, TRACE is the first tennis-video benchmark to make the stroke-level evidence supporting tactical answers an explicit target of both supervision and evaluation.

\begin{figure*}[h]
\centering
\includegraphics[width=0.97\textwidth]{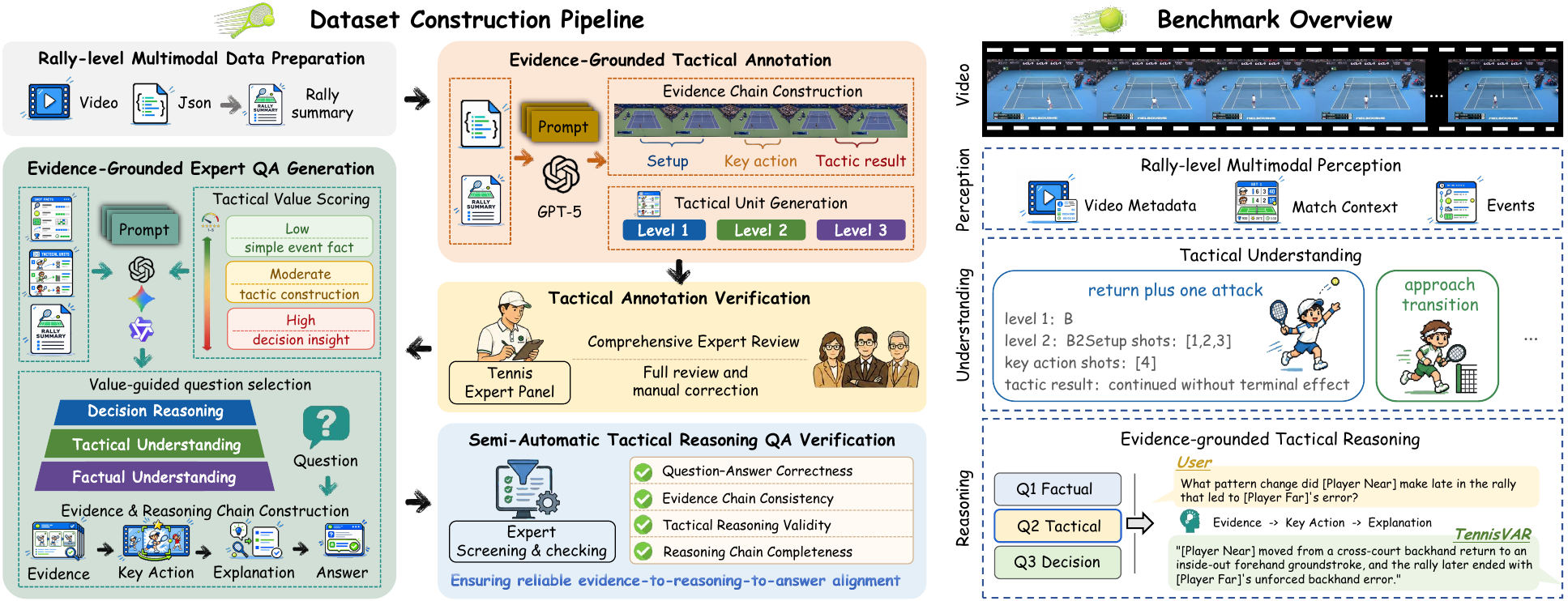}
\caption{\textbf{Construction pipeline and benchmark overview of TRACE.}
Left: Densely annotated strokes are organized into tactical units consisting of setup strokes, key actions, and locally observable outcomes. Expert-verified units are then used to construct evidence-grounded QA instances. Right: TRACE contains three progressive reasoning levels: factual perception, tactical understanding, and decision reasoning. Each rally may contain multiple tactical units but is paired with one QA instance.}
\label{fig:construction_pipeline}
\end{figure*}

\section{Task Formulation and Benchmark Construction}
\label{sec:task_benchmark}

\subsection{Task Formulation}
\label{sec:task_formulation}

Given a rally video $\mathcal V$ with $N$ ordered strokes and a question $q$, the model predicts
\begin{equation}
(a,\mathbf z,\mathcal E,\mathcal K,r)
=f_\theta(\mathcal V,q),
\mathcal K\subseteq\mathcal E\subseteq\{1,\ldots,N\},
\label{eq:trace_task}
\end{equation}
where $a$ is an open-ended answer, $\mathbf z=(z^1,z^2,z^3)$ is a hierarchical tactic label, $\mathcal E$ is the ordered set of supporting strokes, $\mathcal K$ contains the decisive key actions, and $r$ is an evidence-grounded rationale. Each evidence stroke is linked to its racket--ball contact frame, allowing semantic evidence identification and temporal localization to be evaluated jointly.

For benchmark construction, each rally is also organized into question-independent \emph{tactical units}. Each unit specifies a hierarchical tactic, the executing player, setup strokes, key actions, and a locally observable outcome. In contrast, $\mathcal E$ and $\mathcal K$ are question-conditioned and include only the strokes needed to answer $q$.
TRACE organizes questions into three levels:
\textbf{Q1: Factual perception.}
Questions target directly observable stroke or rally facts, such as the hitter, technique, direction, or termination type.
\textbf{Q2: Tactical understanding.}
Questions require reasoning across strokes to identify tactical setups, directional patterns, attack--defense transitions, or offensive--defensive responses.
\textbf{Q3: Decision reasoning.}
Questions examine evidence-supported decisions and their locally observable consequences, such as a player's response to a net approach and the resulting benefit or risk.

\begin{figure*}[t]
\centering
\includegraphics[width=0.99\textwidth]{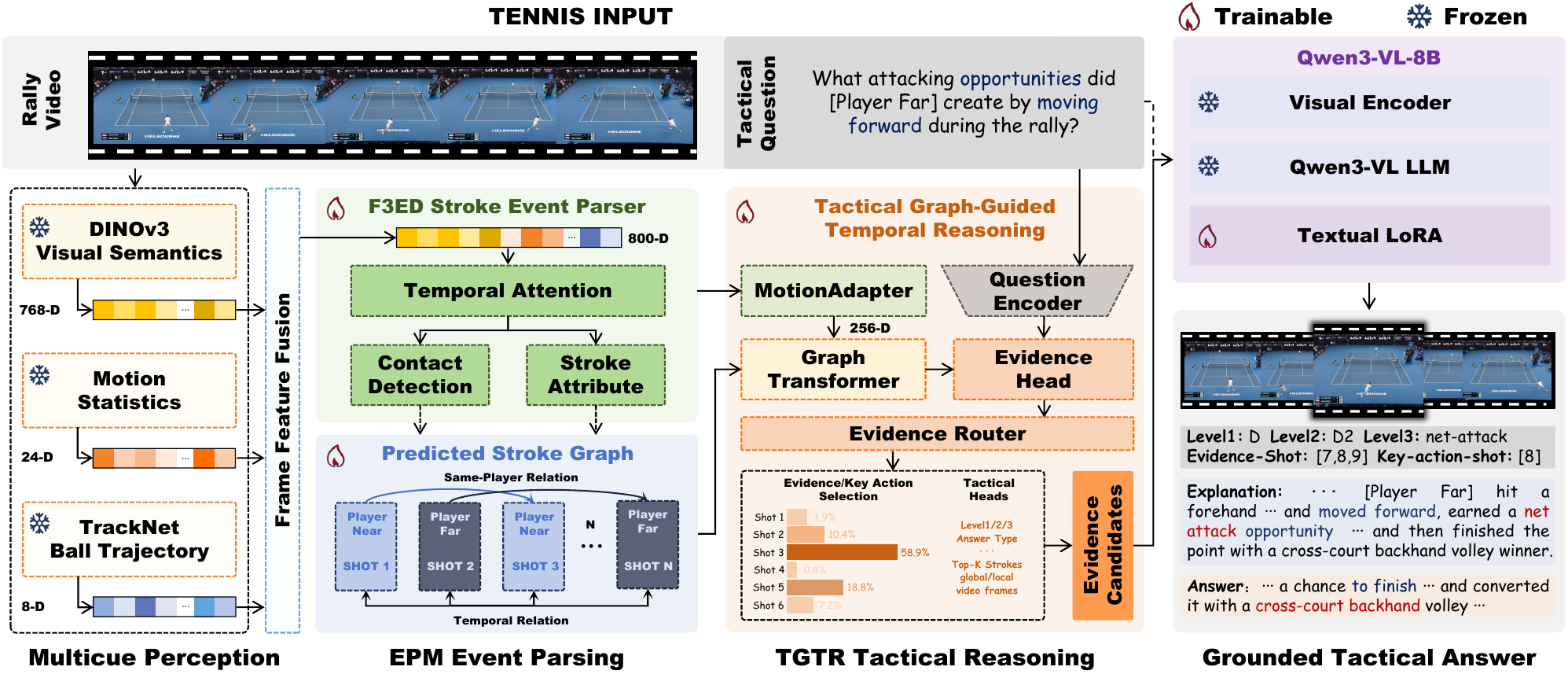}
\caption{\textbf{Overview of TennisVAR.}
Event Parsing reconstructs contact-centered strokes from complementary visual cues. Tactical Reasoning organizes them through temporal and same-player relations, routes question-relevant evidence, and predicts the hierarchical tactic. Selected events and sparse global frames support grounded answer generation.}
\label{fig:TennisVAR_overview}
\end{figure*}

\subsection{TRACE Benchmark Construction}
\label{sec:trace_construction}

TRACE extends the densely timestamped tennis events in F$^3$Set~\cite{liu2025f3set} from event detection to multi-stroke tactical reasoning and evidence-grounded QA. Its source videos cover men's and women's professional matches from Grand Slams, tour-level tournaments, the Olympic Games, and team competitions. As shown in Fig.~\ref{fig:construction_pipeline}, the benchmark is constructed in three stages.

\paragraph{Rally structuring.}
%We convert the original annotations into an ordered sequence of stroke facts, including the hitter, court region, action, technique, direction, forward movement, and outcome. Player identities are replaced with the camera-relative roles \texttt{[Player Near]} and \texttt{[Player Far]}, which remain fixed throughout each clip. Rally length, winner, and termination type are then derived to form a structured rally summary.
We convert the original annotations into an ordered sequence of stroke facts, including the hitter, court region, action, technique, direction, forward movement, and outcome. Player identities are replaced with the camera-relative roles \texttt{[Player Near]} and \texttt{[Player Far]}. Rally length, winner, and termination type are then derived to form a structured rally summary.

\paragraph{Tactical annotation.}
We define a three-level tactical hierarchy containing 6, 17, and 25 classes, respectively, including a \emph{no-primary-tactic} class for rallies without a salient tactical structure. Based on the structured rally facts, a language model proposes candidate tactical units, each describing the corresponding tactical setup, key actions, and locally observable outcome. Before formal annotation, three tennis experts are calibrated using a shared set of examples. Each candidate tactical unit is reviewed by at least two experts, while ambiguous cases are jointly adjudicated. Units involving unsupported intent inference, irrelevant setup strokes, or unobservable outcomes are corrected or discarded.

\paragraph{QA and Evidence annotation.}
To reduce model-specific phrasing bias, we use multiple language-model families to independently generate candidate question--answer pairs from the verified tactical units and rally facts. With the identities of the generating models concealed, human annotators select and rewrite the candidates based on tactical relevance, evidence completeness, and reasoning value. For each retained QA instance, we annotate the minimal supporting-stroke set $\mathcal E$, its decisive key-action subset $\mathcal K$, and a rationale organized as ``tactical setup--key action--observable outcome.'' All fields must remain consistent with the verified rally events. Further details are provided in the supplementary material.

\subsection{Benchmark Scale and Statistics}
% \label{sec:trace_statistics}
% \begin{figure}[t]
% \centering
% \includegraphics[width=0.80\columnwidth]{Figures/Tactical.pdf}
% \caption{\textbf{Hierarchical tactical taxonomy of TRACE.}}
% \label{fig:data}
% \end{figure}
TRACE contains 11,189 rallies from 109 matches featuring 72 players, including 41,485 stroke events, 25,429 expert-verified tactical units, and 11,189 open-ended QA instances. Each rally is paired with one QA instance and contains 2.27 tactical units on average.

The data are split at the source-match level into 7,119 training, 1,805 validation, and 2,265 test rallies, preventing match-specific information from being shared across subsets. The three-level 6/17/25 tactical hierarchy covers serve, return, baseline construction, net transition, and defensive counterattack tactics. The QA dataset contains 3,643 Q1, 6,376 Q2, and 1,170 Q3 instances.

\section{Method}
\label{sec:method}

\subsection{Overview}
\label{sec:overview}

TennisVAR reorganizes a rally from a frame sequence into a question-conditioned tactical event structure. Given $\mathcal V=\{I_t\}_{t=1}^{T}$ and question $q$, its computation is
\begin{equation}
\begin{aligned}
\hat{\mathcal S}=P_{\theta_P}(\mathcal V),
(\hat{\mathbf z},\hat{\mathcal E},\hat{\mathcal K})
=R_{\theta_R}(\hat{\mathcal S},q),
(\hat a,\hat r)=D_{\theta_D}(\mathcal C(q)),
\end{aligned}
\label{eq:event_to_tactic}
\end{equation}
where $\Theta=(\theta_P,\theta_R,\theta_D)$; $\hat{\mathcal S}$ is a predicted stroke-event sequence; $\hat{\mathbf z}$, $\hat{\mathcal E}$, and $\hat{\mathcal K}$ are the tactic, supporting strokes, and key actions; and $\hat a$ and $\hat r$ are the answer and rationale. Evidence contact frames are inherited from $\hat{\mathcal S}$ rather than predicted separately.

As shown in Fig.~\ref{fig:TennisVAR_overview}, the \textbf{Event Parsing Module (EPM)} converts the video into contact-centered semantic strokes, while \textbf{Tactical Graph-Guided Temporal Reasoning (TGTR)} models their dependencies and routes the question-relevant action chain into tactical prediction. The language model only verbalizes this evidence-bearing structure.

\subsection{Event Parsing Module}
\label{sec:eventization}

Racket--ball contacts are brief but tactically decisive. To preserve them, we fuse appearance, short-term motion, and ball-trajectory cues:
\begin{equation}
\mathbf{x}_t=\phi_{\mathrm{fuse}}\!\left([\mathbf{x}^{\mathrm{app}}_t;\mathbf{x}^{\mathrm{mot}}_t;\mathbf{x}^{\mathrm{ball}}_t]\right).
\label{eq:fused_observation}
\end{equation}
DINOv3~\cite{simeoni2026dinov3} captures players and court context, the motion stream captures abrupt changes, and TrackNet~\cite{huang2019tracknet} provides the ball trajectory.

A local-to-global F3ED encoder~\cite{liu2025f3set} localizes contacts and predicts observable stroke attributes. Temporal decoding produces
\begin{equation}
\hat{\mathcal S}=\{s_i\}_{i=1}^{\hat N},\qquad s_i=(\tau_i,\mathbf x_{\tau_i},\boldsymbol{\eta}_i),
\label{eq:event_output}
\end{equation}
where $\tau_i$ is the contact frame, $\mathbf x_{\tau_i}$ is its fused visual feature, and $\boldsymbol{\eta}_i$ contains the hitter, stroke type, direction, and outcome. This sequence is the shared interface between visual perception and tactical reasoning.
We train contact detection with continuous-target focal binary cross-entropy~\cite{lin2017focal} and supervise attributes only at annotated contacts:
\begin{equation}
\mathcal L_{\mathrm{evt}}=\mathcal L_{\mathrm{det}}+\lambda_{\mathrm{attr}}\mathcal L_{\mathrm{attr}}.
\label{eq:event_loss}
\end{equation}

The EPM therefore establishes the contact-aligned semantic units on which all subsequent relations and evidence predictions are defined.

\subsection{Tactical Graph-Guided Temporal Reasoner}
\label{sec:tgtr}

A tactic emerges from dependencies among strokes rather than from any stroke in isolation. TGTR captures both the chronological exchange and each player's action transitions across intervening returns.

\paragraph{Relational event structure.}
Each stroke forms a node whose token combines its visual feature, timestamp, rally position, and attributes. We construct
\begin{equation}
\mathcal G=(\mathcal V_s,\mathcal R_{\mathrm{time}}\cup\mathcal R_{\mathrm{player}}),
\label{eq:stroke_graph}
\end{equation}
where $\mathcal R_{\mathrm{time}}$ links adjacent strokes and $\mathcal R_{\mathrm{player}}$ links successive actions by the same predicted player across an intervening return. Relation-conditioned message passing~\cite{schlichtkrull2018rgcn} followed by a Transformer~\cite{vaswani2017attention} yields contextualized stroke tokens $\{\mathbf g_i\}$ and a rally representation $\mathbf g_{\mathcal G}$. All nodes and relations come from EPM predictions.

\paragraph{Evidence-routed tactical inference.}
Because a rally may contain several tactical patterns, two question-conditioned heads score whether stroke $i$ supports the answer ($h=E$) or is a key action ($h=K$):
\begin{equation}
u_i^h=f_h([\mathbf g_i;\mathbf q]),\qquad p_i^h=\sigma(u_i^h),\qquad h\in\{E,K\}.
\label{eq:grounding_scores}
\end{equation}
Let $\mathbf p^h=(p_i^h)_{i=1}^{\hat N}$. The selected evidence is ordered by contact time, key actions are restricted to this set, and $\hat{\mathcal F}=(\tau_i)_{i\in\hat{\mathcal E}}$. With $\alpha_i=\operatorname{softmax}_i(u_i^E)$, the Evidence Router forms
\begin{equation}
\mathbf g_R=\operatorname{Fuse}\!\left(\mathbf g_{\mathcal G},\sum_i\alpha_i\mathbf g_i,\mathbf q\right).
\label{eq:evidence_router}
\end{equation}
The three tactic heads operate on $\mathbf g_R$, so evidence participates in tactical prediction rather than being attached afterward.

\paragraph{Learning in the predicted event space.}
To avoid oracle stroke indices, we align annotated evidence frames $\mathcal F$ with predicted contacts $\hat{\mathcal T}=\{\tau_i\}$ through maximum-cardinality, minimum-offset one-to-one matching:
\begin{equation}
\pi^\star=\arg\min_{\pi\in\Pi_\delta^{\max}}\sum_{(f,\tau)\in\pi}|f-\tau|.
\label{eq:evidence_alignment}
\end{equation}
Here $\Pi_\delta^{\max}$ contains admissible maximum-cardinality matchings within temporal tolerance $\delta$. Matched events receive evidence and Key-action labels. Let $\mathcal L_{\mathrm{tac}}=\sum_{\ell=1}^{3}\lambda_\ell\mathcal L_\ell$. TGTR is optimized by
\begin{equation}
\mathcal L_{\mathrm{TGTR}}=\lambda_E\mathcal L_E+\lambda_K\mathcal L_K+\mathcal L_{\mathrm{tac}}+\lambda_S\mathcal L_S.
\label{eq:tgtr_loss}
\end{equation}
Here $\mathcal L_E$ and $\mathcal L_K$ are grounding losses, $\mathcal L_\ell$ supervises tactic level $\ell$, and $\mathcal L_S$ is an auxiliary semantic loss. Training in the predicted event space reduces the gap between training and inference.

\paragraph{Answer realization and inference.}
\label{sec:generation}

The language model receives sparse global frames, local windows around selected contacts, and a serialized event table:
\begin{equation}
\mathcal C(q)=[q;\mathcal V_g;\mathcal V_l(\hat{\mathcal E});\mathcal T_c].
\label{eq:generation_context}
\end{equation}
The table $\mathcal T_c$ contains event identifiers, timestamps, attributes, and grounding scores. Qwen3-VL~\cite{bai2025qwen3vl} generates the answer and rationale, while the tactic, evidence, and key-action fields come from TGTR. At inference, all events, relations, and evidence are predicted from $(\mathcal V,q)$; no oracle input is used.

\section{Experiments}
\label{sec:experiment}

\begin{table*}[t]
    \centering
    %\footnotesize
    \setlength{\tabcolsep}{1.8pt}
    \renewcommand{\arraystretch}{1.10}

    \begin{tabular}{@{}ll*{11}{c}@{}}
        \toprule
        \multirow{2}{*}{\textbf{Setting}}
        & \multirow{2}{*}{\textbf{Model}}
        & \multicolumn{5}{c}{\textbf{Evidence}}
        & \multicolumn{2}{c}{\textbf{Tactical}}
        & \multicolumn{3}{c}{\textbf{Text}}
        & \multirow{2}{*}{\textbf{Total}} \\
        \cmidrule(lr){3-7}
        \cmidrule(lr){8-9}
        \cmidrule(lr){10-12}
        &
        & \shortstack{T-F1@8}
        & \shortstack{T-F1@16}
        & \shortstack{T-IoU@4}
        & \shortstack{F-Acc@8}
        & \shortstack{F-Acc@16}
        & \shortstack{Hier.F1}
        & \shortstack{Key Acc.}
        & R-L
        & CIDEr
        & B-4
        & \\
        \midrule

\multirow[c]{5}{*}{%
    \rotatebox[origin=c]{90}{%
        \shortstack[c]{Open-source\\Zero-shot}%
    }%
}
        & Llama-4-Scout
        & 0.09 & 0.09 & 0.00 & 0.00 & 0.00
        & 5.52 & 0.00 & 19.81 & 1.88 & 7.08 & 2.76 \\

        & Llama-4-Maverick
        & 0.92 & 1.28 & 0.89 & 0.00 & 0.00
        & 6.24 & 2.47 & 19.41 & 2.25 & 7.82 & 3.58 \\

        & DeepSeek-VL2
        & 2.97 & 4.23 & 1.60 & 0.00 & 0.53
        & 1.51 & 3.50 & 7.38 & 0.51 & 3.19 & 2.42 \\

        & Qwen2.5-VL-3B
        & 5.92 & 10.69 & 3.95 & 0.44 & 0.84
        & 2.93 & 10.07 & 0.63 & 0.03 & 0.20 & 4.19 \\

        & Qwen2.5-VL-7B
        & 10.62 & 16.55 & 9.88 & 1.96 & 3.89
        & 7.73 & 20.91 & 10.45 & 0.74 & 3.64 & 9.57 \\

        & Qwen3-VL-8B
        & 17.94 & 21.47 & 10.87 & 1.50 & 3.99
        & 8.15 & 22.96 & 19.75 & 2.20 & 6.78 & 12.16 \\
        \midrule

\multirow[c]{5}{*}{%
    \rotatebox[origin=c]{90}{%
        \shortstack[c]{Closed-source\\Zero-shot}%
    }%
}
        & Gemini-3-Pro
        & 30.15 & 47.53 & 12.65 & 2.13 & 6.42
        & 11.50 & 24.11 & 26.83 & 3.13 & 9.98 & 17.89 \\

        & Gemini-3.1-Pro
        & 30.59 & 48.36 & 12.41 & 3.54 & 8.55
        & 13.33 & 26.02 & 26.66 & 3.05 & 9.63 & 18.87 \\

        & Claude-Sonnet-4.6
        & 27.95 & 38.20 & 19.50 & 0.00 & 0.00
        & 12.51 & 16.28 & 22.15 & 0.28 & 5.80 & 14.77 \\

        & Claude-Opus-4.6
        & 27.09 & 36.21 & 18.23 & 0.00 & 0.00
        & 14.59 & 16.59 & 22.47 & 0.39 & 5.92 & 14.75 \\

        & GPT-5.5
        & 37.03 & 49.49 & \underline{24.54} & 1.31 & 2.57
        & 12.67 & 16.50 & 26.33 & 2.16 & 9.00 & 18.36 \\
        \midrule

\multirow[c]{3}{*}{\rotatebox[origin=c]{90}{SFT}}
        & InternVL3-8B
        & \underline{53.10}
        & \underline{68.84}
        & 23.16
        & \underline{27.42}
        & \underline{43.89}
        & 59.91
        & 44.75
        & 55.28
        & 23.82
        & 32.99
        & 44.81 \\

        & Qwen2.5-VL-7B
        & 45.93
        & 66.60
        & 19.61
        & 21.41
        & 35.58
        & \underline{64.90}
        & 45.76
        & 54.05
        & 22.31
        & 31.60
        & 42.71 \\

        & Qwen3-VL-8B
        & 49.46
        & 67.96
        & 21.06
        & 25.52
        & 39.56
        & 64.49
        & \underline{47.13}
        & \underline{56.20}
        & \underline{25.32}
        & \underline{34.45}
        & \underline{44.83} \\
        \midrule

        \multicolumn{2}{@{}l}{\textbf{TennisVAR (Ours)}}
        & \textbf{73.04}
        & \textbf{76.59}
        & \textbf{56.19}
        & \textbf{47.86}
        & \textbf{51.66}
        & \textbf{70.98}
        & \textbf{52.27}
        & \textbf{57.98}
        & \textbf{27.12}
        & \textbf{36.28}
        & \textbf{57.11} \\
        \bottomrule
    \end{tabular}
        \caption{
        Comparison with representative zero-shot and supervised fine-tuning
        baselines. Higher is better for all metrics. Best and second-best
        results are shown in bold and underlined, respectively.
    }
    \label{tab:main_results}
\end{table*}

\subsection{Experimental Setup}
\label{sec:setting}
TennisVAR uses Qwen3-VL-8B~\cite{bai2025qwen3vl} as its vision-language generator. The EPM employs an F3ED encoder that combines DINOv3 appearance features, short-term motion features, and TrackNet ball-trajectory features, followed by local temporal modules and a temporal Transformer. It is trained for 40 epochs with AdamW~\cite{loshchilov2019decoupled}, a batch size of 64, and a learning rate of $2\times10^{-4}$.
TGTR is trained for 120 epochs with AdamW, a batch size of 64, and a learning rate of $1.5\times10^{-3}$. We set $\lambda_E=\lambda_K=2.0$, $\lambda_1=\lambda_2=\lambda_3=1.0$, and $\lambda_S=0.5$.
For answer generation, all pretrained Qwen3-VL parameters are frozen and rank-32 LoRA modules~\cite{hu2022lora} are optimized for 5 epochs with a learning rate of $2\times10^{-5}$ and an effective batch size of 32. Training uses $8\times$NVIDIA H20 96GB GPUs.

\subsection{Main Results}
\label{sec:main_results}

\paragraph{Evaluation metrics.} We evaluate the model from three aspects: \textbf{evidence localization}, \textbf{understanding}, and \textbf{linguistic quality}.
For \textbf{evidence localization}, we use Temporal F1@8, Temporal F1@16, Temporal IoU@4, Frame Accuracy@8, and Frame Accuracy@16 to measure how well the predicted evidence strokes match the reference evidence under different temporal and frame-level criteria.
For \textbf{understanding}, we use Hierarchical Tactic F1 and Key-action Accuracy. Hierarchical Tactic F1 evaluates tactical predictions under a hierarchical taxonomy, while Key-action Accuracy measures whether the decisive stroke in a rally is correctly identified.
For \textbf{linguistic quality}, we use BLEU-4~\cite{papineni2002bleu}, ROUGE-L~\cite{lin2004rouge}, and CIDEr~\cite{vedantam2015cider}. Together, these metrics assess whether the model truly understands tactical information and grounds its answers in video evidence, rather than relying only on language patterns or response templates.
The \textbf{Total} score is computed by first averaging the metrics within each group and then combining the evidence, tactical, and language groups with weights of 0.50, 0.30, and 0.20, respectively. All metrics are reported on a 0--100 scale.

\paragraph{Baselines.}
We compare TennisVAR with zero-shot open-weight and proprietary MLLMs. Open-weight models include Llama-4-Scout and Llama-4-Maverick~\cite{meta2025llama4}, DeepSeek-VL2~\cite{wu2024deepseek}, Qwen2.5-VL-3B/7B~\cite{bai2025qwen25vl}, and Qwen3-VL-8B~\cite{bai2025qwen3vl}. Proprietary models include Gemini-3-Pro and Gemini-3.1-Pro~\cite{googledeepmind2025gemini3pro,googledeepmind2026gemini31pro}, Claude-Opus-4.6 and Claude-Sonnet-4.6~\cite{anthropic2026opus46,anthropic2026sonnet46}, and GPT-5.5~\cite{openai2026gpt55}.
We additionally fine-tune InternVL3-8B~\cite{zhu2025internvl3}, Qwen2.5-VL-7B, and Qwen3-VL-8B on TRACE. 
All baseline models are evaluated on the same test set.
%All baselines receive the same raw video and question and are evaluated on the same test set.

\paragraph{Overall comparison.} As shown in Table~\ref{tab:main_results}, TennisVAR achieves the best performance across all ten component metrics as well as the overall score, demonstrating strong and balanced capabilities in evidence localization, tactical understanding, and answer generation. Zero-shot MLLMs can often produce plausible responses, yet remain substantially less effective at identifying the stroke-level evidence that supports them. For example, GPT-5.5 obtains a Temporal F1@8 of 37.03 and a Temporal IoU@4 of 24.54, whereas TennisVAR achieves 73.04 and 56.19, respectively. This contrast suggests that general-purpose models may draw on language priors and coarse global video context, but have difficulty precisely grounding their answers in the relevant strokes.

TennisVAR also substantially outperforms the supervised fine-tuning baselines. Compared with the strongest SFT baselines on the corresponding metrics, TennisVAR improves Temporal F1@8, Temporal F1@16, Temporal IoU@4, Frame Accuracy@8, and Frame Accuracy@16 by 19.94, 7.75, 33.03, 20.44, and 7.77 percentage points, respectively. The particularly large gain in Temporal IoU@4 indicates that TennisVAR not only retrieves relevant segments of a rally but also aligns the predicted evidence more precisely with the underlying stroke events.

The model also delivers consistent improvements in tactical reasoning. TennisVAR achieves a Hierarchical Tactic F1 of 70.98 and a Key-action Accuracy of 52.27, exceeding the strongest SFT baselines by 6.08 and 5.14 percentage points, respectively. These results are consistent with the intended role of question-conditioned graph reasoning: modeling dependencies across strokes helps the model identify tactically decisive stages of a rally, rather than inferring tactical labels primarily from isolated local observations.

TennisVAR consistently improves all conventional text-generation metrics, outperforming Qwen3-VL-8B by 1.78, 1.80, and 1.83 percentage points on ROUGE-L, CIDEr, and BLEU-4, respectively. Notably, several general-purpose models achieve competitive text-similarity scores despite substantially weaker evidence localization. This observation suggests that lexical-overlap-based metrics primarily capture surface-level agreement with reference answers and may not fully reflect whether an answer is supported by the correct strokes. We therefore evaluate answer quality jointly with stroke-level evidence localization, providing a more comprehensive assessment of evidence-grounded tactical reasoning.

\begin{table*}[htbp]
    \centering
    % \resizebox{\textwidth}{!}{
    \setlength{\tabcolsep}{2.6pt}
    \renewcommand{\arraystretch}{1.10}
    {
    \begin{tabular}{lccccccccccc}
        \toprule
        Setting
        & T-F1@8
        & T-F1@16
        & T-IoU@4
        & F-Acc@8
        & F-Acc@16
        & Hier.\ F1
        & Key Acc.
        & R-L
        & CIDEr
        & B-4
        & Total \\
        \midrule

        Video-only
        & 49.46 & 67.96 & 21.06 & 25.52 & 39.56
        & 64.49 & 47.13 & 56.20 & 25.32 & 34.45
        & 44.83 \\

        w/o EPM
        & 61.58 & 63.52 & 40.73 & 34.53 & 40.53
        & 68.89 & 45.14 & 57.45 & 26.41 & 35.61
        & 49.16 \\

        w/o DINOv3
        & 59.91 & 70.29 & 43.70 & 38.41 & 46.98
        & 69.76 & 44.06 & 57.66 & 26.66 & 35.78
        & 51.01 \\

        w/o TrackNet
        & 62.13 & 73.62 & 55.70 & 47.77 & 50.79
        & 69.08 & 50.33 & 57.32 & 26.31 & 35.30
        & 54.84 \\

        w/o Motion
        & 64.22 & 74.17 & 54.91 & 46.70 & 49.80
        & 67.56 & 49.32 & 57.46 & 26.33 & 35.46
        & 54.46 \\

        w/o TGTR
        & 55.90 & 68.72 & 41.57 & 32.32 & 44.26
        & 66.70 & 41.35 & 55.62 & 24.27 & 33.47
        & 48.04 \\

        w/o Evidence Router
        & 63.56 & 71.24 & 54.60 & 44.68 & 49.05
        & 60.80 & 46.45 & 56.91 & 25.98 & 35.04
        & 52.26 \\

        \midrule
        \textbf{TennisVAR (Ours)}
        & \textbf{73.04}
        & \textbf{76.59}
        & \textbf{56.19}
        & \textbf{47.86}
        & \textbf{51.66}
        & \textbf{70.98}
        & \textbf{52.27}
        & \textbf{57.98}
        & \textbf{27.12}
        & \textbf{36.28}
        & \textbf{57.11} \\

        \bottomrule
    \end{tabular}
    }
    % }
    \caption{
        Component ablation results.
        The best result in each column is highlighted in bold.
    }
    \label{tab:ablation_results}
\end{table*}

\begin{table}[t]
\centering
%\footnotesize
\setlength{\tabcolsep}{2.8pt}
\renewcommand{\arraystretch}{1.08}

\begin{tabular}{@{}llccc@{}}
    \toprule
    \textbf{Level}
    & \textbf{Model}
    & \shortstack{\textbf{T-F1}\textbf{@8}}
    & \shortstack{\textbf{T-IoU}\textbf{@4}}
    & \shortstack{\textbf{Hier.}\textbf{F1}} \\
    \midrule

    \multirow[c]{4}{*}{Q1}
    & Qwen3-VL-8B-SFT
    & 50.33 & 22.49 & 66.62 \\

    & Qwen2.5-VL-7B-SFT
    & 44.37 & 21.45 & 65.22 \\

    & InternVL3-8B-SFT
    & 59.84 & 27.63 & 63.45 \\

    & \textbf{TennisVAR}
    & \textbf{83.50}
    & \textbf{70.35}
    & \textbf{73.17} \\
    \midrule

    \multirow[c]{4}{*}{Q2}
    & Qwen3-VL-8B-SFT
    & 51.06 & 21.56 & 63.19 \\

    & Qwen2.5-VL-7B-SFT
    & 48.73 & 20.06 & 64.91 \\

    & InternVL3-8B-SFT
    & 52.04 & 21.98 & 55.40 \\

    & \textbf{TennisVAR}
    & \textbf{71.08}
    & \textbf{52.54}
    & \textbf{69.54} \\
    \midrule

    \multirow[c]{4}{*}{Q3}
    & Qwen3-VL-8B-SFT
    & 37.36 & 13.94 & 60.48 \\

    & Qwen2.5-VL-7B-SFT
    & 33.56 & 11.65 & 60.14 \\

    & InternVL3-8B-SFT
    & 40.17 & 17.43 & 51.59 \\

    & \textbf{TennisVAR}
    & \textbf{54.88}
    & \textbf{37.64}
    & \textbf{66.38} \\
    \bottomrule
\end{tabular}
\caption{
Performance across reasoning levels. 
The best result within each level is shown in bold.
}
\label{tab:reasoning_levels}
\end{table}

\paragraph{Performance across reasoning levels.}

Table~\ref{tab:reasoning_levels} compares TennisVAR with the three SFT baselines on factual perception (Q1), tactical understanding (Q2), and decision reasoning (Q3).
TennisVAR consistently achieves the best performance at all three levels. 
On Q1, it surpasses the strongest baseline by 23.66 points in T-F1@8, 42.72 points in T-IoU@4, and 6.55 points in Hierarchical F1. On Q2, the corresponding gains are 19.04, 30.56, and 4.63 points.
%On Q3, TennisVAR continues to lead all metrics, but its evidence-localization scores decrease to 54.88 and 37.64. This trend confirms that recovering evidence for decision-level reasoning becomes more challenging as the required tactical dependencies grow more complex.
Although Q3 is the most demanding level, TennisVAR retains clear margins of 14.71 T-F1@8, 20.21 T-IoU@4, and 5.90 Hierarchical Tactic F1 over the strongest SFT baselines, demonstrating that its advantage persists as tactical dependencies become more complex.

\subsection{Qualitative Analysis}
\label{sec:qualitative}

\begin{figure}[t]
\centering
\includegraphics[width=\columnwidth]{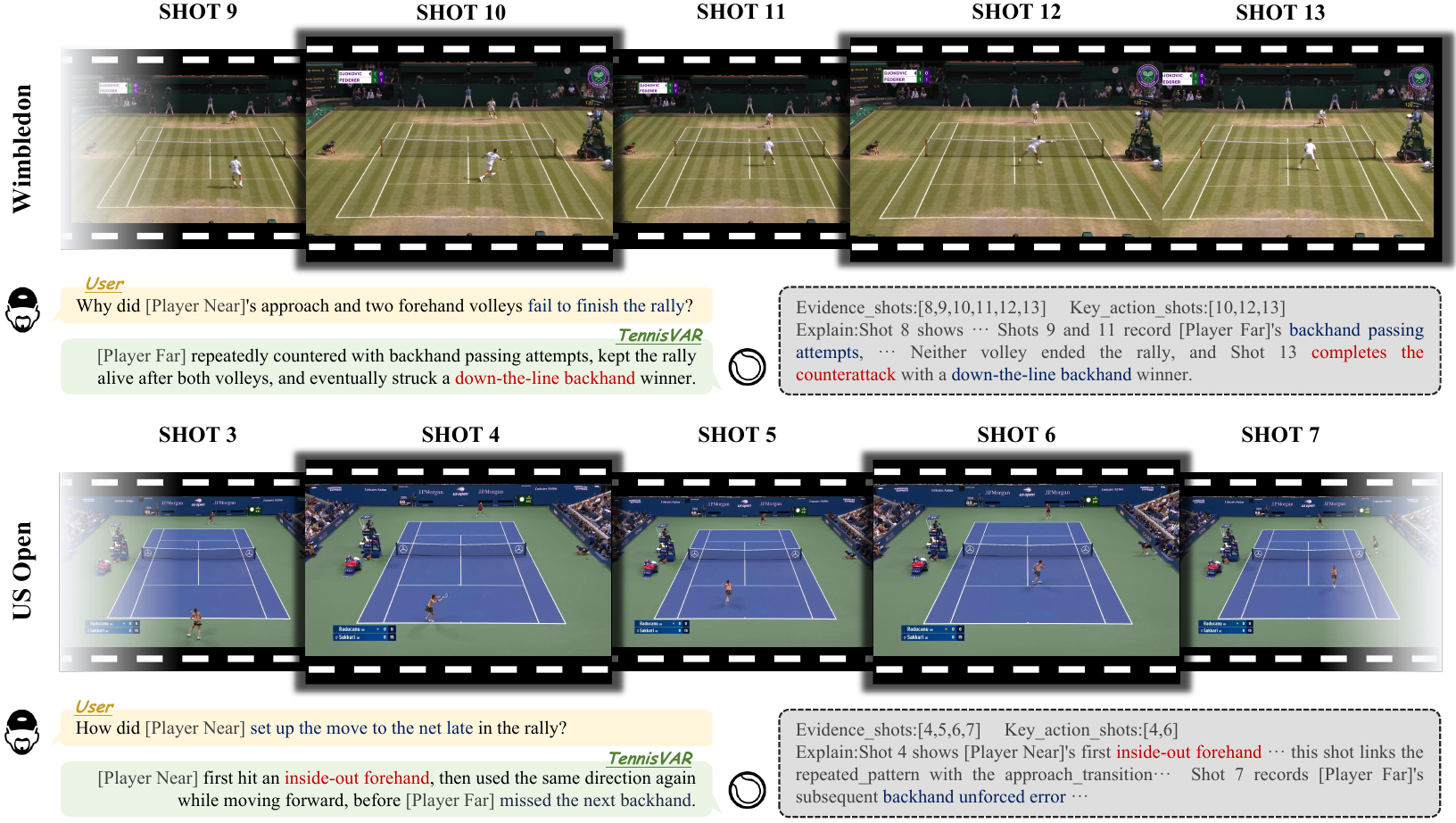}
\caption{\textbf{Qualitative examples of evidence-grounded tactical reasoning.} TennisVAR links temporally distributed strokes to its answer and rationale.}
\label{fig:qualitative_examples}
\end{figure}

Figure~\ref{fig:qualitative_examples} shows how TennisVAR grounds tactical answers in ordered stroke evidence. In the Wimbledon example, it connects repeated backhand passing attempts with the final down-the-line winner to explain why two volleys failed to finish the rally. In the US Open example, it links repeated inside-out forehands and forward movement to the subsequent net approach. These cases illustrate how the EPM recovers contact-aligned strokes and Tactical Reasoning organizes them into question-relevant evidence chains.

\subsection{Ablation Studies}
\label{sec:ablation}

\paragraph{Ablation settings.}
Table~\ref{tab:ablation_results} evaluates the two core designs. \emph{Video-only} removes both EPM and TGTR. \emph{w/o TGTR} retains parsed events but removes Tactical Reasoning, while \emph{w/o Evidence Router} retains relational reasoning without question-conditioned evidence routing. For the EPM, we further remove appearance, trajectory, and motion cues individually. All variants share the same generator and training protocol.

\paragraph{Event Parsing Module.}
% Adding the EPM to the video-only model, while leaving TGTR removed, improves Temporal F1@8 by 6.44 points and Temporal IoU@4 by 20.51 points. This gain shows that contact-centered semantic events provide a substantially stronger grounding space than raw frames. Removing DINOv3 causes the largest overall drop of 6.09 points, while removing TrackNet and motion reduces Temporal F1@8 by 10.91 and 8.82 points, respectively. Appearance supplies the main semantic context, whereas trajectory and motion preserve complementary contact cues.
Adding EPM to the video-only model without TGTR improves Temporal F1@8 by 6.44 points and Temporal IoU@4 by 20.51 points, indicating that contact-centered semantic events provide stronger temporal grounding than raw frames. Conversely, removing EPM from the full model reduces these metrics by 11.46 and 15.46 points, respectively. Among the EPM inputs, removing DINOv3 causes the largest overall performance drop of 6.10 points, while removing TrackNet or motion decreases Temporal F1@8 by 10.91 and 8.82 points. These results suggest that appearance provides the primary semantic context, while trajectory and motion offer complementary contact cues.

\paragraph{Tactical Reasoning.}
Removing TGTR from the full model decreases Total by 9.07 points, Temporal F1@8 by 17.14 points, Temporal IoU@4 by 14.62 points, and Key-action Accuracy by 10.92 points. Within TGTR, removing the Evidence Router lowers Hierarchical Tactic F1 by 10.18 points and Total by 4.85 points. These results confirm that relational event modeling recovers cross-stroke tactical structure, while evidence routing connects that structure to the question-specific tactic.

\section{Conclusion}
\label{sec:conclusion}

We introduced \textbf{stroke-evidence-grounded tactical reasoning}, a new rally-level task that evaluates both tactical predictions and the stroke events supporting them. To support this task, we constructed \textbf{TRACE}, a large-scale expert-annotated benchmark that unifies fine-grained stroke events, cross-stroke tactical relations, hierarchical tactics, and ordered evidence attribution. We further proposed \textbf{TennisVAR}, an evidence-grounded MLLM following an ``event--relation--evidence--tactic'' paradigm. By parsing explicit stroke events and modeling rally progression and same-player decision dependencies, TennisVAR substantially improves evidence localization, key-action identification, and tactical prediction. These results demonstrate the importance of structured event reasoning and explicit evidence grounding for reliable tennis-video understanding.

\bibliography{aaai2027}

@inproceedings{zhang2023videollama,
  author       = {Hang Zhang and
                  Xin Li and
                  Lidong Bing},
  editor       = {Yansong Feng and
                  Els Lefever},
  title        = {Video-LLaMA: An Instruction-tuned Audio-Visual Language Model for
                  Video Understanding},
  booktitle    = {Proceedings of the 2023 Conference on Empirical Methods in Natural
                  Language Processing, {EMNLP} 2023 - System Demonstrations, Singapore,
                  December 6-10, 2023},
  pages        = {543--553},
  publisher    = {Association for Computational Linguistics},
  year         = {2023},
  url          = {https://doi.org/10.18653/v1/2023.emnlp-demo.49},
  doi          = {10.18653/V1/2023.EMNLP-DEMO.49}
}

@inproceedings{maaz2024videochatgpt,
  author       = {Muhammad Maaz and
                  Hanoona Abdul Rasheed and
                  Salman Khan and
                  Fahad Khan},
  editor       = {Lun{-}Wei Ku and
                  Andre Martins and
                  Vivek Srikumar},
  title        = {Video-ChatGPT: Towards Detailed Video Understanding via Large Vision
                  and Language Models},
  booktitle    = {Proceedings of the 62nd Annual Meeting of the Association for Computational
                  Linguistics (Volume 1: Long Papers), {ACL} 2024, Bangkok, Thailand,
                  August 11-16, 2024},
  pages        = {12585--12602},
  publisher    = {Association for Computational Linguistics},
  year         = {2024},
  url          = {https://doi.org/10.18653/v1/2024.acl-long.679},
  doi          = {10.18653/V1/2024.ACL-LONG.679}
}

@misc{bai2025qwen3vl,
  title={Qwen3-vl technical report},
  author={Bai, Shuai and Cai, Yuxuan and Chen, Ruizhe and Chen, Keqin and Chen, Xionghui and Cheng, Zesen and Deng, Lianghao and Ding, Wei and Gao, Chang and Ge, Chunjiang and others},
  journal={arXiv preprint arXiv:2511.21631},
  year={2025}
}

@inproceedings{deliege2021soccernet,
  author       = {Adrien Deli{\`{e}}ge and
                  Anthony Cioppa and
                  Silvio Giancola and
                  Meisam Jamshidi Seikavandi and
                  Jacob V. Dueholm and
                  Kamal Nasrollahi and
                  Bernard Ghanem and
                  Thomas B. Moeslund and
                  Marc Van Droogenbroeck},
  title        = {SoccerNet-v2: {A} Dataset and Benchmarks for Holistic Understanding
                  of Broadcast Soccer Videos},
  booktitle    = {{IEEE} Conference on Computer Vision and Pattern Recognition Workshops,
                  {CVPR} Workshops 2021, virtual, June 19-25, 2021},
  pages        = {4508--4519},
  publisher    = {Computer Vision Foundation / {IEEE}},
  year         = {2021},
  url          = {https://openaccess.thecvf.com/content/CVPR2021W/CVSports/html/Deliege\_SoccerNet-v2\_A\_Dataset\_and\_Benchmarks\_for\_Holistic\_Understanding\_of\_Broadcast\_CVPRW\_2021\_paper.html},
  doi          = {10.1109/CVPRW53098.2021.00508}
}

@inproceedings{shao2020finegym,
  author       = {Dian Shao and
                  Yue Zhao and
                  Bo Dai and
                  Dahua Lin},
  title        = {FineGym: {A} Hierarchical Video Dataset for Fine-Grained Action Understanding},
  booktitle    = {2020 {IEEE/CVF} Conference on Computer Vision and Pattern Recognition,
                  {CVPR} 2020, Seattle, WA, USA, June 13-19, 2020},
  pages        = {2613--2622},
  publisher    = {Computer Vision Foundation / {IEEE}},
  year         = {2020},
  url          = {https://openaccess.thecvf.com/content\_CVPR\_2020/html/Shao\_FineGym\_A\_Hierarchical\_Video\_Dataset\_for\_Fine-Grained\_Action\_Understanding\_CVPR\_2020\_paper.html},
  doi          = {10.1109/CVPR42600.2020.00269}
}

@inproceedings{xu2022finediving,
  author       = {Jinglin Xu and
                  Yongming Rao and
                  Xumin Yu and
                  Guangyi Chen and
                  Jie Zhou and
                  Jiwen Lu},
  title        = {FineDiving: {A} Fine-grained Dataset for Procedure-aware Action Quality
                  Assessment},
  booktitle    = {{IEEE/CVF} Conference on Computer Vision and Pattern Recognition,
                  {CVPR} 2022, New Orleans, LA, USA, June 18-24, 2022},
  pages        = {2939--2948},
  publisher    = {{IEEE}},
  year         = {2022},
  url          = {https://doi.org/10.1109/CVPR52688.2022.00296},
  doi          = {10.1109/CVPR52688.2022.00296}
}

@inproceedings{huang2019tracknet,
  author       = {Yu{-}Chuan Huang and
                  I{-}No Liao and
                  Ching{-}Hsuan Chen and
                  Ts{\`{\i}}{-}U{\'{\i}} Ik and
                  Wen{-}Chih Peng},
  title        = {TrackNet: {A} Deep Learning Network for Tracking High-speed and Tiny
                  Objects in Sports Applications},
  booktitle    = {16th {IEEE} International Conference on Advanced Video and Signal
                  Based Surveillance, {AVSS} 2019, Taipei, Taiwan, September 18-21,
                  2019},
  pages        = {1--8},
  publisher    = {{IEEE}},
  year         = {2019},
  url          = {https://doi.org/10.1109/AVSS.2019.8909871},
  doi          = {10.1109/AVSS.2019.8909871}
}

@inproceedings{liu2025f3set,
  author       = {Zhaoyu Liu and
                  Kan Jiang and
                  Murong Ma and
                  Zhe Hou and
                  Yun Lin and
                  Jin Song Dong},
  title        = {F3Set: Towards Analyzing Fast, Frequent, and Fine-grained Events from
                  Videos},
  booktitle    = {The Thirteenth International Conference on Learning Representations,
                  {ICLR} 2025, Singapore, April 24-28, 2025},
  publisher    = {OpenReview.net},
  year         = {2025},
  url          = {https://openreview.net/forum?id=vlg5WRKHxh}
}

@inproceedings{fu2025videomme,
  author       = {Chaoyou Fu and
                  Yuhan Dai and
                  Yongdong Luo and
                  Lei Li and
                  Shuhuai Ren and
                  Renrui Zhang and
                  Zihan Wang and
                  Chenyu Zhou and
                  Yunhang Shen and
                  Mengdan Zhang and
                  Peixian Chen and
                  Yanwei Li and
                  Shaohui Lin and
                  Sirui Zhao and
                  Ke Li and
                  Tong Xu and
                  Xiawu Zheng and
                  Enhong Chen and
                  Caifeng Shan and
                  Ran He and
                  Xing Sun},
  title        = {Video-MME: The First-Ever Comprehensive Evaluation Benchmark of Multi-modal
                  LLMs in Video Analysis},
  booktitle    = {{IEEE/CVF} Conference on Computer Vision and Pattern Recognition,
                  {CVPR} 2025, Nashville, TN, USA, June 11-15, 2025},
  pages        = {24108--24118},
  publisher    = {Computer Vision Foundation / {IEEE}},
  year         = {2025},
  url          = {https://openaccess.thecvf.com/content/CVPR2025/html/Fu\_Video-MME\_The\_First-Ever\_Comprehensive\_Evaluation\_Benchmark\_of\_Multi-modal\_LLMs\_in\_CVPR\_2025\_paper.html},
  doi          = {10.1109/CVPR52734.2025.02245}
}

@inproceedings{wu2024longvideobench,
  author       = {Haoning Wu and
                  Dongxu Li and
                  Bei Chen and
                  Junnan Li},
  editor       = {Amir Globersons and
                  Lester Mackey and
                  Danielle Belgrave and
                  Angela Fan and
                  Ulrich Paquet and
                  Jakub M. Tomczak and
                  Cheng Zhang},
  title        = {LongVideoBench: {A} Benchmark for Long-context Interleaved Video-Language
                  Understanding},
  booktitle    = {Advances in Neural Information Processing Systems 37: Annual Conference
                  on Neural Information Processing Systems 2024, NeurIPS 2024, Vancouver,
                  BC, Canada, December 10 - 15, 2024},
  year         = {2024},
  url          = {http://papers.nips.cc/paper\_files/paper/2024/hash/329ad516cf7a6ac306f29882e9c77558-Abstract-Datasets\_and\_Benchmarks\_Track.html}
}

@inproceedings{zhou2025glimpse,
  author       = {Yiyang Zhou and
                  Linjie Li and
                  Shi Qiu and
                  Zhengyuan Yang and
                  Yuyang Zhao and
                  Siwei Han and
                  Yangfan He and
                  Kangqi Li and
                  Haonian Ji and
                  Zihao Zhao and
                  Haibo Tong and
                  Lijuan Wang and
                  Huaxiu Yao},
  editor       = {Christos Christodoulopoulos and
                  Tanmoy Chakraborty and
                  Carolyn Rose and
                  Violet Peng},
  title        = {{GLIMPSE:} Do Large Vision-Language Models Truly Think With Videos
                  or Just Glimpse at Them?},
  booktitle    = {Proceedings of the 2025 Conference on Empirical Methods in Natural
                  Language Processing, {EMNLP} 2025, Suzhou, China, November 4-9, 2025},
  pages        = {27842--27856},
  publisher    = {Association for Computational Linguistics},
  year         = {2025},
  url          = {https://doi.org/10.18653/v1/2025.emnlp-main.1415},
  doi          = {10.18653/V1/2025.EMNLP-MAIN.1415}
  
}

@inproceedings{xiao2021nextqa,
  author       = {Junbin Xiao and
                  Xindi Shang and
                  Angela Yao and
                  Tat{-}Seng Chua},
  title        = {NExT-QA: Next Phase of Question-Answering to Explaining Temporal Actions},
  booktitle    = {{IEEE} Conference on Computer Vision and Pattern Recognition, {CVPR}
                  2021, virtual, June 19-25, 2021},
  pages        = {9777--9786},
  publisher    = {Computer Vision Foundation / {IEEE}},
  year         = {2021},
  url          = {https://openaccess.thecvf.com/content/CVPR2021/html/Xiao\_NExT-QA\_Next\_Phase\_of\_Question-Answering\_to\_Explaining\_Temporal\_Actions\_CVPR\_2021\_paper.html},
  doi          = {10.1109/CVPR46437.2021.00965}
}

@inproceedings{xiao2024nextgqa,
  author       = {Junbin Xiao and
                  Angela Yao and
                  Yicong Li and
                  Tat{-}Seng Chua},
  title        = {Can {I} Trust Your Answer? Visually Grounded Video Question Answering},
  booktitle    = {{IEEE/CVF} Conference on Computer Vision and Pattern Recognition,
                  {CVPR} 2024, Seattle, WA, USA, June 16-22, 2024},
  pages        = {13204--13214},
  publisher    = {{IEEE}},
  year         = {2024},
  url          = {https://doi.org/10.1109/CVPR52733.2024.01254},
  doi          = {10.1109/CVPR52733.2024.01254}
}

@article{li2026sportsqa,
  author       = {Haopeng Li and
                  Andong Deng and
                  Jun Liu and
                  Hossein Rahmani and
                  Yulan Guo and
                  Bernt Schiele and
                  Mohammed Bennamoun and
                  Qiuhong Ke},
  title        = {Sports-QA: {A} Large-Scale Video Question Answering Benchmark for
                  Complex and Professional Sports},
  journal      = {Int. J. Comput. Vis.},
  volume       = {134},
  number       = {5},
  pages        = {196},
  year         = {2026},
  url          = {https://doi.org/10.1007/s11263-026-02734-1},
  doi          = {10.1007/S11263-026-02734-1}
}

@inproceedings{xia2025sportu,
  author       = {Haotian Xia and
                  Zhengbang Yang and
                  Junbo Zou and
                  Rhys Tracy and
                  Yuqing Wang and
                  Chi Lu and
                  Christopher Lai and
                  Yanjun He and
                  Xun Shao and
                  Zhuoqing Xie and
                  Yuan{-}Fang Wang and
                  Weining Shen and
                  Hanjie Chen},
  title        = {{SPORTU:} {A} Comprehensive Sports Understanding Benchmark for Multimodal
                  Large Language Models},
  booktitle    = {The Thirteenth International Conference on Learning Representations,
                  {ICLR} 2025, Singapore, April 24-28, 2025},
  publisher    = {OpenReview.net},
  year         = {2025},
  url          = {https://openreview.net/forum?id=x1yOHtFfDh},
}

@inproceedings{rao2025unisoccer,
  author       = {Jiayuan Rao and
                  Haoning Wu and
                  Hao Jiang and
                  Ya Zhang and
                  Yanfeng Wang and
                  Weidi Xie},
  title        = {Towards Universal Soccer Video Understanding},
  booktitle    = {{IEEE/CVF} Conference on Computer Vision and Pattern Recognition,
                  {CVPR} 2025, Nashville, TN, USA, June 11-15, 2025},
  pages        = {8384--8394},
  publisher    = {Computer Vision Foundation / {IEEE}},
  year         = {2025},
  url          = {https://openaccess.thecvf.com/content/CVPR2025/html/Rao\_Towards\_Universal\_Soccer\_Video\_Understanding\_CVPR\_2025\_paper.html},
  doi          = {10.1109/CVPR52734.2025.00785}
}

@inproceedings{xia2026sportr,
title={SportR: A Benchmark for Multimodal Large Language Model Reasoning in Sports},
author={Haotian Xia and Haonan Ge and Junbo Zou and Hyun Woo Choi and Xuebin Zhang and Danny Suradja and Botao Rui and Ethan Tran and Wendy Jin and Zhen Ye and Xiyang Lin and Christopher Lai and Shengjie Zhang and Junwen Miao and Shichao Chen and Rhys Tracy and Vicente Ordonez and Weining Shen and Hanjie Chen},
booktitle={The Fourteenth International Conference on Learning Representations},
year={2026},
url={https://openreview.net/forum?id=cPCGB402ff}
}

@misc{bao2025tennistv,
  title={TennisTV: Do Multimodal Large Language Models Understand Tennis Rallies?},
  author={Bao, Zhongyuan and Zhang, Lejun},
  journal={arXiv preprint arXiv:2509.15602},
  year={2025}
}

@misc{liu2026tennisexpert,
  title={TennisExpert: towards expert-level analytical sports video understanding},
  author={Liu, Zhaoyu and Weng, Xi and Hu, Lianyu and Hou, Zhe and Jiang, Kan and Dong, Jin Song and Liu, Yang},
  journal={arXiv preprint arXiv:2603.13397},
  year={2026},
  volume={abs/2603.13397},
  bibsource    = {dblp computer science bibliography, https://dblp.org}
}

@inproceedings{rao2024matchtime,
  author       = {Jiayuan Rao and
                  Haoning Wu and
                  Chang Liu and
                  Yanfeng Wang and
                  Weidi Xie},
  editor       = {Yaser Al{-}Onaizan and
                  Mohit Bansal and
                  Yun{-}Nung Chen},
  title        = {MatchTime: Towards Automatic Soccer Game Commentary Generation},
  booktitle    = {Proceedings of the 2024 Conference on Empirical Methods in Natural
                  Language Processing, {EMNLP} 2024, Miami, FL, USA, November 12-16,
                  2024},
  pages        = {1671--1685},
  publisher    = {Association for Computational Linguistics},
  year         = {2024},
  url          = {https://doi.org/10.18653/v1/2024.emnlp-main.99},
  doi          = {10.18653/V1/2024.EMNLP-MAIN.99}
}

@article{simeoni2026dinov3,
  author       = {Oriane Sim{\'{e}}oni and
                  Huy V. Vo and
                  Maximilian Seitzer and
                  Federico Baldassarre and
                  Maxime Oquab and
                  Cijo Jose and
                  Vasil Khalidov and
                  Marc Szafraniec and
                  Seung Eun Yi and
                  Micha{\"{e}}l Ramamonjisoa and
                  Francisco Massa and
                  Daniel Haziza and
                  Luca Wehrstedt and
                  Jianyuan Wang and
                  Timoth{\'{e}}e Darcet and
                  Th{\'{e}}o Moutakanni and
                  Leonel Sentana and
                  Claire Roberts and
                  Andrea Vedaldi and
                  Jamie Tolan and
                  John Brandt and
                  Camille Couprie and
                  Julien Mairal and
                  Herv{\'{e}} J{\'{e}}gou and
                  Patrick Labatut and
                  Piotr Bojanowski},
  title        = {DINOv3},
  journal      = {Trans. Mach. Learn. Res.},
  volume       = {2026},
  year         = {2026},
  url          = {https://openreview.net/forum?id=2NlGyqNjns},
}

@inproceedings{lin2017focal,
  author       = {Tsung{-}Yi Lin and
                  Priya Goyal and
                  Ross B. Girshick and
                  Kaiming He and
                  Piotr Doll{\'{a}}r},
  title        = {Focal Loss for Dense Object Detection},
  booktitle    = {{IEEE} International Conference on Computer Vision, {ICCV} 2017, Venice,
                  Italy, October 22-29, 2017},
  pages        = {2999--3007},
  publisher    = {{IEEE} Computer Society},
  year         = {2017},
  url          = {https://doi.org/10.1109/ICCV.2017.324},
  doi          = {10.1109/ICCV.2017.324}
}

@inproceedings{schlichtkrull2018rgcn,
  author       = {Michael Sejr Schlichtkrull and
                  Thomas N. Kipf and
                  Peter Bloem and
                  Rianne van den Berg and
                  Ivan Titov and
                  Max Welling},
  editor       = {Aldo Gangemi and
                  Roberto Navigli and
                  Maria{-}Esther Vidal and
                  Pascal Hitzler and
                  Rapha{\"{e}}l Troncy and
                  Laura Hollink and
                  Anna Tordai and
                  Mehwish Alam},
  title        = {Modeling Relational Data with Graph Convolutional Networks},
  booktitle    = {The Semantic Web - 15th International Conference, {ESWC} 2018, Heraklion,
                  Crete, Greece, June 3-7, 2018, Proceedings},
  series       = {Lecture Notes in Computer Science},
  volume       = {10843},
  pages        = {593--607},
  publisher    = {Springer},
  year         = {2018},
  url          = {https://doi.org/10.1007/978-3-319-93417-4\_38},
  doi          = {10.1007/978-3-319-93417-4\_38}
}

@inproceedings{vaswani2017attention,
  author       = {Ashish Vaswani and
                  Noam Shazeer and
                  Niki Parmar and
                  Jakob Uszkoreit and
                  Llion Jones and
                  Aidan N. Gomez and
                  Lukasz Kaiser and
                  Illia Polosukhin},
  editor       = {Isabelle Guyon and
                  Ulrike von Luxburg and
                  Samy Bengio and
                  Hanna M. Wallach and
                  Rob Fergus and
                  S. V. N. Vishwanathan and
                  Roman Garnett},
  title        = {Attention is All you Need},
  booktitle    = {Advances in Neural Information Processing Systems 30: Annual Conference
                  on Neural Information Processing Systems 2017, December 4-9, 2017,
                  Long Beach, CA, {USA}},
  pages        = {5998--6008},
  year         = {2017},
  url          = {https://proceedings.neurips.cc/paper/2017/hash/3f5ee243547dee91fbd053c1c4a845aa-Abstract.html}
}

@inproceedings{papineni2002bleu,
  author       = {Kishore Papineni and
                  Salim Roukos and
                  Todd Ward and
                  Wei{-}Jing Zhu},
  title        = {Bleu: a Method for Automatic Evaluation of Machine Translation},
  booktitle    = {Proceedings of the 40th Annual Meeting of the Association for Computational
                  Linguistics, July 6-12, 2002, Philadelphia, PA, {USA}},
  pages        = {311--318},
  publisher    = {{ACL}},
  year         = {2002},
  url          = {https://aclanthology.org/P02-1040/},
  doi          = {10.3115/1073083.1073135}
}

@inproceedings{lin2004rouge,
  author    = {Chin-Yew Lin},
  title     = {{ROUGE}: A Package for Automatic Evaluation of Summaries},
  booktitle = {Text Summarization Branches Out},
  pages     = {74--81},
  publisher = {Association for Computational Linguistics},
  year      = {2004}
}

@inproceedings{vedantam2015cider,
  author       = {Ramakrishna Vedantam and
                  C. Lawrence Zitnick and
                  Devi Parikh},
  title        = {CIDEr: Consensus-based image description evaluation},
  booktitle    = {{IEEE} Conference on Computer Vision and Pattern Recognition, {CVPR}
                  2015, Boston, MA, USA, June 7-12, 2015},
  pages        = {4566--4575},
  publisher    = {{IEEE} Computer Society},
  year         = {2015},
  url          = {https://doi.org/10.1109/CVPR.2015.7299087},
  doi          = {10.1109/CVPR.2015.7299087}
}

@inproceedings{zhu2026mmrv,
title={{MMR}-V: What's Left Unsaid? A Benchmark for Multimodal Deep Reasoning in Videos},
author={Kejian Zhu and Zhuoran Jin and Hongbang Yuan and Jiachun Li and Shangqing Tu and Pengfei Cao and Yubo Chen and Kang Liu and Jun Zhao},
booktitle={The Fourteenth International Conference on Learning Representations},
year={2026},
url={https://openreview.net/forum?id=xk8EqWDPQw}
}

@misc{zhang2026cast,
  title={CaST-Bench: Benchmarking Causal Chain-Grounded Spatio-Temporal Reasoning for Video Question Answering},
  author={Zhang, Mingfang and Pan, Jingjing and Kumar, Ashutosh and Saini, Rajat and Erdogan, Mustafa and Yang, Hsuan-Kung and Kang, Caixin and Huang, Yifei and Sato, Yoichi and Kong, Quan},
  journal={arXiv preprint arXiv:2605.23216},
  year={2026}
}

@inproceedings{loshchilov2019decoupled,
  author       = {Ilya Loshchilov and
                  Frank Hutter},
  title        = {Decoupled Weight Decay Regularization},
  booktitle    = {7th International Conference on Learning Representations, {ICLR} 2019,
                  New Orleans, LA, USA, May 6-9, 2019},
  publisher    = {OpenReview.net},
  year         = {2019},
  url          = {https://openreview.net/forum?id=Bkg6RiCqY7}
}

@inproceedings{hu2022lora,
  author       = {Edward J. Hu and
                  Yelong Shen and
                  Phillip Wallis and
                  Zeyuan Allen{-}Zhu and
                  Yuanzhi Li and
                  Shean Wang and
                  Lu Wang and
                  Weizhu Chen},
  title        = {LoRA: Low-Rank Adaptation of Large Language Models},
  booktitle    = {The Tenth International Conference on Learning Representations, {ICLR}
                  2022, Virtual Event, April 25-29, 2022},
  publisher    = {OpenReview.net},
  year         = {2022},
  url          = {https://openreview.net/forum?id=nZeVKeeFYf9},
}

@misc{bai2025qwen25vl,
      title={Qwen2.5-VL Technical Report}, 
      author={Shuai Bai and Keqin Chen and Xuejing Liu and Jialin Wang and Wenbin Ge and Sibo Song and Kai Dang and Peng Wang and Shijie Wang and Jun Tang and Humen Zhong and Yuanzhi Zhu and Mingkun Yang and Zhaohai Li and Jianqiang Wan and Pengfei Wang and Wei Ding and Zheren Fu and Yiheng Xu and Jiabo Ye and Xi Zhang and Tianbao Xie and Zesen Cheng and Hang Zhang and Zhibo Yang and Haiyang Xu and Junyang Lin},
      year={2025},
      journal={arXiv preprint arXiv:2502.13923}
}

@misc{zhu2025internvl3,
  title={Internvl3: Exploring advanced training and test-time recipes for open-source multimodal models},
  author={Zhu, Jinguo and Wang, Weiyun and Chen, Zhe and Liu, Zhaoyang and Ye, Shenglong and Gu, Lixin and Tian, Hao and Duan, Yuchen and Su, Weijie and Shao, Jie and others},
  journal={arXiv preprint arXiv:2504.10479},
  year={2025}
}

@misc{meta2025llama4,
  author       = {{Meta AI}},
  title        = {The {Llama 4} Herd: The Beginning of a New Era of Natively Multimodal {AI} Innovation},
  year         = {2025},
  howpublished = {\url{https://ai.meta.com/blog/llama-4-multimodal-intelligence/}},
  note         = {Accessed: 2026-07-23}
}

@misc{wu2024deepseek,
  title={Deepseek-vl2: Mixture-of-experts vision-language models for advanced multimodal understanding},
  author       = {Zhiyu Wu and
                  Xiaokang Chen and
                  Zizheng Pan and
                  Xingchao Liu and
                  Wen Liu and
                  Damai Dai and
                  Huazuo Gao and
                  Yiyang Ma and
                  Chengyue Wu and
                  Bingxuan Wang and
                  Zhenda Xie and
                  Yu Wu and
                  Kai Hu and
                  Jiawei Wang and
                  Yaofeng Sun and
                  Yukun Li and
                  Yishi Piao and
                  Kang Guan and
                  Aixin Liu and
                  Xin Xie and
                  Yuxiang You and
                  Kai Dong and
                  Xingkai Yu and
                  Haowei Zhang and
                  Liang Zhao and
                  Yisong Wang and
                  Chong Ruan},
  journal={arXiv preprint arXiv:2412.10302},
  year={2024}
}

@misc{googledeepmind2025gemini3pro,
  author       = {{Google DeepMind}},
  title        = {{Gemini 3 Pro} Model Card},
  year         = {2025},
  howpublished = {\url{https://deepmind.google/models/model-cards/gemini-3-pro/}},
  note         = {Model released November 2025; model card accessed: 2026-07-23}
}

@misc{googledeepmind2026gemini31pro,
  author       = {{Google DeepMind}},
  title        = {{Gemini 3.1 Pro} Model Card},
  year         = {2026},
  howpublished = {\url{https://deepmind.google/models/model-cards/gemini-3-1-pro/}},
  note         = {Published: 2026-02-19; accessed: 2026-07-23}
}

@misc{anthropic2026sonnet46,
  author       = {{Anthropic}},
  title        = {Introducing {Claude Sonnet 4.6}},
  year         = {2026},
  howpublished = {\url{https://www.anthropic.com/news/claude-sonnet-4-6}},
  note         = {Published: 2026-02-17; accessed: 2026-07-23}
}

@misc{anthropic2026opus46,
  author       = {{Anthropic}},
  title        = {Introducing {Claude Opus 4.6}},
  year         = {2026},
  howpublished = {\url{https://www.anthropic.com/news/claude-opus-4-6}},
  note         = {Published: 2026-02-05; accessed: 2026-07-23}
}

@misc{openai2026gpt55,
  author       = {{OpenAI}},
  title        = {{GPT-5.5} System Card},
  year         = {2026},
  howpublished = {\url{https://openai.com/index/gpt-5-5-system-card/}},
  note         = {Published: 2026-04-23; accessed: 2026-07-28}
}


\end{document}